\documentclass[10pt,twocolumn,letterpaper]{article}

\usepackage{wacv}
\usepackage{times}
\usepackage{epsfig}
\usepackage{graphicx}
\usepackage{amsmath}
\usepackage{amssymb}

\usepackage{booktabs}
\usepackage{multirow}

\def\wacvPaperID{1194} 

\wacvfinalcopy 

\ifwacvfinal
\def\assignedStartPage{1} 
\fi

\ifwacvfinal
\usepackage[breaklinks=true,bookmarks=false]{hyperref}
\else
\usepackage[pagebackref=true,breaklinks=true,colorlinks,bookmarks=false]{hyperref}
\fi

\ifwacvfinal
\else
\fi

\begin{document}

\title{RefVOS: A Closer Look at Referring Expressions for Video Object Segmentation}








\author{Miriam Bellver$^1$,  Carles Ventura$^2$,  Carina Silberer$^{3, 4}$, \\ Ioannis Kazakos$^5$, Jordi Torres$^1$ and Xavier Giro-i-Nieto$^6$ \vspace{4mm} \\ 
$^1$Barcelona Supercomputing Center  \:\:\:
$^2$Universitat Oberta de Catalunya  \:\:\: \\
$^3$Institute for NLP, University of Stuttgart \:\:\: 
$^4$Universitat Pompeu Fabra \:\:\:  \\
$^5$National Technical University of Athens \:\:\: 
$^6$Universitat Politècnica de Catalunya \:\:\:
}

\maketitle

\begin{abstract}
The task of video object segmentation with referring expressions (language-guided VOS) is to, given a linguistic phrase and a video, generate binary masks for the object to which the phrase refers. Our work argues that existing benchmarks used for this task are mainly composed of trivial cases, in which referents can be identified with simple phrases. Our analysis relies on a new categorization of the phrases in the DAVIS-2017 and Actor-Action datasets into trivial and non-trivial REs, with the non-trivial REs annotated with seven RE semantic categories. We leverage this data to analyze the results of RefVOS, a novel neural network that obtains competitive results for the task of language-guided image segmentation and state of the art results for language-guided VOS. Our study indicates that the major challenges for the task are related to understanding motion and static actions.
\end{abstract}

\section{Introduction}
\label{sec:introduction}



\begin{figure*}[!tbp]
  \centering
  \begin{minipage}[t]{0.384\textwidth}

    \includegraphics[width=\columnwidth]{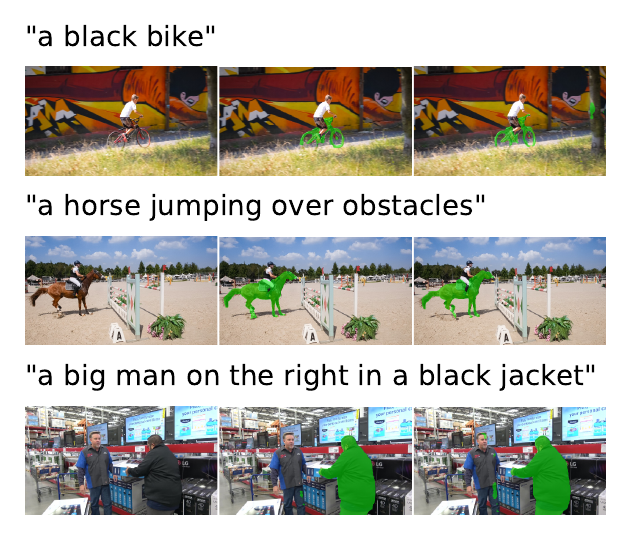}
    \label{fig:refcoco_best_testA}
  \end{minipage}
\begin{minipage}[t]{0.3264\textwidth}

    \includegraphics[width=\columnwidth]{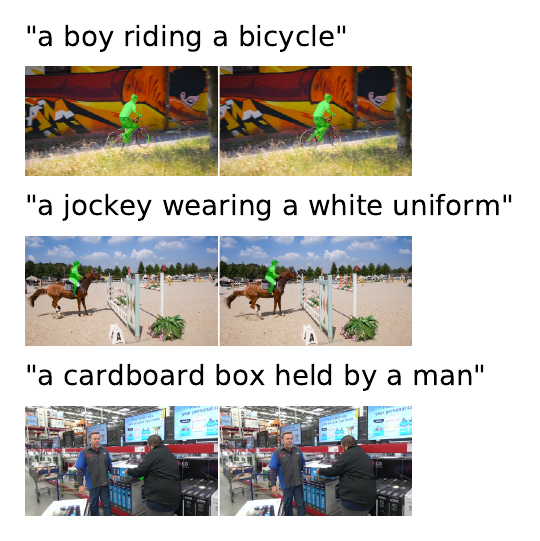}
   
  \end{minipage}
  \vspace*{-\baselineskip}
  \caption{Video sequences for DAVIS 2017 with language queries and our results. The first column shows a reference frame, the second to third columns depict the masks produced by our model when given the language query shown on top. Finally, the fourth to fifth columns show the results for the language query shown on top of these columns, which refers to another object of the video sequence.
  }
  \label{fig:davis_best_full_video}
\end{figure*}

Video Object Segmentation~(VOS)~\cite{perazzi2016benchmark,xu2018youtube} has been traditionally considered on setups in which the user would manually annotate a frame in the video, and a segmentation system would generate a pixel-wise binary mask for the object in all video frames where it is visible. 
Our work aims at improving the human computer interaction by allowing linguistic expressions as initialization cues, instead of interactive segmentations under the form of a detailed binary mask, bounding box, scribble or point. 
In particular, we focus on referring expressions (REs) that allow the identification of an individual object in a discourse or scene (the \textsl{referent}). 
For instance, Figure \ref{fig:davis_best_full_video} depicts REs related to one of the objects contained in a video sequence, which is highlighted in green.

Language-guided Video Object Segmentation (LVOS) was first addressed Khoreva~\etal~\cite{khoreva2018video}, and tackled later by Gavrilyuk~\etal~\cite{gavrilyuk2018actor} and Wang~\etal~\cite{wang2019asymmetric}. 
Compared to related works on still images~\cite{yu2018mattnet,chen2019see}, REs for video objects may be more complex, as they can refer to variations in the properties of the objects, such as a change of location or appearance.
The particularities of REs for videos were initially addressed by Khoreva~\etal~\cite{khoreva2018video}, who built a dataset of REs divided in two categories: REs for the first frame of a video, and REs for the full clip.
Our work proposes another approach for analyzing the performance of the state of the art in VOS with REs. We identify seven categories of REs and use them to annotate existing datasets.

We address both the language-guided image segmentation and the language-guided video object segmentation tasks with \textit{RefVOS}, 
our end-to-end deep neural network that 
leverages BERT language representations \cite{devlin2019bert} to encode the phrases.
RefVOS achieves results comparable to the state of the art for the RefCOCO dataset of still images~\cite{kazemzadeh2014referitgame}, and improves the state of the art over the DAVIS-2017~\cite{pont20172017} and Actor-Action datasets (A2D)~\cite{xu2015can} for video with the phrases collected by Khoreva~\etal~\cite{khoreva2018video} and Gavrilyuk~\etal~\cite{gavrilyuk2018actor}, respectively.
We also identify the categories of REs which are most challenging for RefVOS.


Our main contributions are summarized as follows: (1) an end-to-end model, \textit{RefVOS}, that achieves state of the art performance with available phrases for DAVIS-2017 and A2D benchmarks, (2) a novel categorization of REs tailored to the video scenario with an analysis of the current benchmarks, and (3) an extension of A2D with additional REs with varying semantic information to analyze the limitations and strengths of our model according to the proposed linguistic categories.

The models, code and extended dataset of REs are available in \href{https://github.com/miriambellver/refvos}{\textcolor{blue}{https://github.com/miriambellver/refvos}}.







\section{Related Work}
\label{sec:related}

\noindent
\textbf{Language-guided Image Segmentation:} The task, also known as referring image segmentation, was first tackled by Hu~\etal~\cite{hu2016segmentation}. They use VGG-16~\cite{simonyan2014very} to obtain a visual representation of the image, and a Long-Short Term Memory~(LSTM) network to obtain an embedding of the RE. From the concatenation of visual and language features, the segmentation of the referred object is obtained. Posterior work \cite{liu2017recurrent} explored how to include multi-scale semantics in the pipeline, by proposing a Recurrent Refinement Network that takes pyramidal features and refines the segmentation masks progressively. Liu~\etal~\cite{liu2019learning} argued to better represent the multi-modality of the task by jointly modeling the language and the image with a multi-modal LSTM that encodes the sequential interactions between words, visual features and the spatial information. With the same purpose of better capturing the multi-modal nature of this task, long-range correlations between the visual and language representations can be reinforced by learning a cross-modal attention module (CMSA)~\cite{ye2019cross}. Building on the same idea, BRINet~\cite{hu2020bi} adds a gated bi-directional fusion module to better integrate multi-level features. Another relevant work is STEP~\cite{chen2019see}. They present a method that learns a visual-textual co-embedding, and that iteratively refines the textual embedding of the RE with a Convolutional Recurrent Neural Network in a collaborative learning setup to improve the segmentation.
An alternative consists in using off-the-shelf object detectors, like MAttNet~\cite{yu2018mattnet}. In this case, a language attention network decomposes REs into three components: subject, location, and relationships, and merges the features obtained for each into single phrase embeddings. Given the object candidate by an off-the-shelf object detector model and a RE, the visual module dynamically weights scores from all three modules to fuse them. A different approach proposed recently is CMPC~\cite{huang2020referring}, which leverages multi-modal graph reasoning to identify the target objects.





RefVOS is a simpler model trained end-to-end that obtains a performance comparable to the state of the art on still images.


\noindent
\textbf{Language-guided Video Object Tracking:} Object Tracking is a similar task to Video Object Segmentation as it also follows a referent across video frames, but in the tracking case the model localizes the object with a bounding box instead of a binary mask. Li~\etal~\cite{li2017tracking} and Feng~\etal~\cite{feng2020real} tackle the object tracking problem given a linguistic phrase instead of using the bounding box at the first frame.

Our work provides pixel-wise segmentation masks that could be easily converted into bounding boxes, and at the same time avoid the annotation ambiguities present when bounding boxes overlap.

\noindent
\textbf{Language-guided Video Object Segmentation~(LVOS):} VOS \cite{perazzi2016benchmark,xu2018youtube} has traditionally focused on semi-supervised setups in which
a binary mask of the object is provided for the first frame of the video.
Khoreva~\etal~\cite{khoreva2018video} propose to replace the mask supervision with a linguistic expression.
In their work, they extend the  DAVIS-2017 dataset~\cite{pont20172017}, a popular dataset for VOS, by collecting referring expressions for the annotated objects. They provide two different kinds of annotations from two annotators each: \textit{first frame} annotations are the ones that are produced by only looking at the first frame of the video, whereas \textit{full video} annotations are produced after seeing the whole video sequence.
They use the image-based MAttNet~\cite{yu2018mattnet} model pretrained on RefCOCO to ground the localization of the referred object, and then train a segmentation network with DAVIS-2017 to produce the pixel-wise prediction. 
Temporal consistency is enforced, so that bounding boxes are coherent across frames, with a post-processing step.
To the authors' knowledge, Khoreva~\etal~\cite{khoreva2018video} is the only work previous to ours that focuses on REs for video object segmentation. Related work by Gavrilyuk~\etal~\cite{gavrilyuk2018actor} addresses a similar task by segmenting video objects given a natural language query. They extend the Actor-Action Dataset~(A2D)~\cite{xu2015can} by collecting phrases, but some of them may be ambiguous with respect to the intended referent, as they were not produced with the aim of reference, but description. 
The authors propose a model with a 3D convolutional encoder and dynamic filters that specialize to localize the referred objects. Wang~\etal~\cite{wang2019asymmetric} also leverage 3D convolutional networks, adding cross-attention between the visual and the language encoder. Concurrent to our work, Seo~\etal~\cite{seourvos} propose URVOS, a model for LVOS composed of a cross-modal attention module for the visual and lingual features, and a memory attention module to leverage information from past predictions in a sequence. 

Our work proposes a simpler model trained end-to-end that treats each video frame independently and outperforms all previous works.

\noindent
\textbf{Referring Expression Categorization:}
RefCOCO, RefCOCO+~\cite{yu2016modeling} and RefCOCOg \cite{mao2015generation} are datasets that provide REs for the still images in MSCOCO~\cite{lin2014mscoco}. 
The datasets focus on different aspects related to the difficulty of REs: 
the REs for RefCOCO and RefCOCO+ were collected using the interactive ReferIt two-player game~\cite{kazemzadeh2014referitgame}, designed to crowdsource expressions that uniquely identify the target referents. 
However, for RefCOCO+, \textit{location} information was disallowed. 
RefCOCOg, in turn, collected  non-interactively, only contains \textit{non-trivial} instances of target objects, that is, there is at least one other object in an image of the same class. 
The CLEVR dataset~\cite{johnson2017clevr} contains objects of certain shapes, attributes such as sizes and colors, and spatial relationships. 
CLEVR uses synthetic images and phrases designed to test VQA systems, while our work focuses on human-produced language and natural videos. 

Khoreva~\etal~\cite{khoreva2018video} categorize the REs they collected for DAVIS-2017 in order to analyze the effectiveness of their proposed model. 
This is similar to our work, however, while they distinguish  REs according to their length  and whether they contain spatial words (e.g.,~\textit{left}) or verbs, we propose a more fine-grained, semantic categorization that also distinguishes between different aspects of verb meaning related to motion and object relations. 
Khoreva~\etal~\cite{khoreva2018video}  furthermore analyze the REs in DAVIS-2017 with respect to the parts of speech they contain, while we use our \textit{semantic} categories for dataset analysis.

\section{Model}
\label{sec:single-frame-models}
\begin{figure*}
  \centering
  \includegraphics[width=0.75\textwidth]{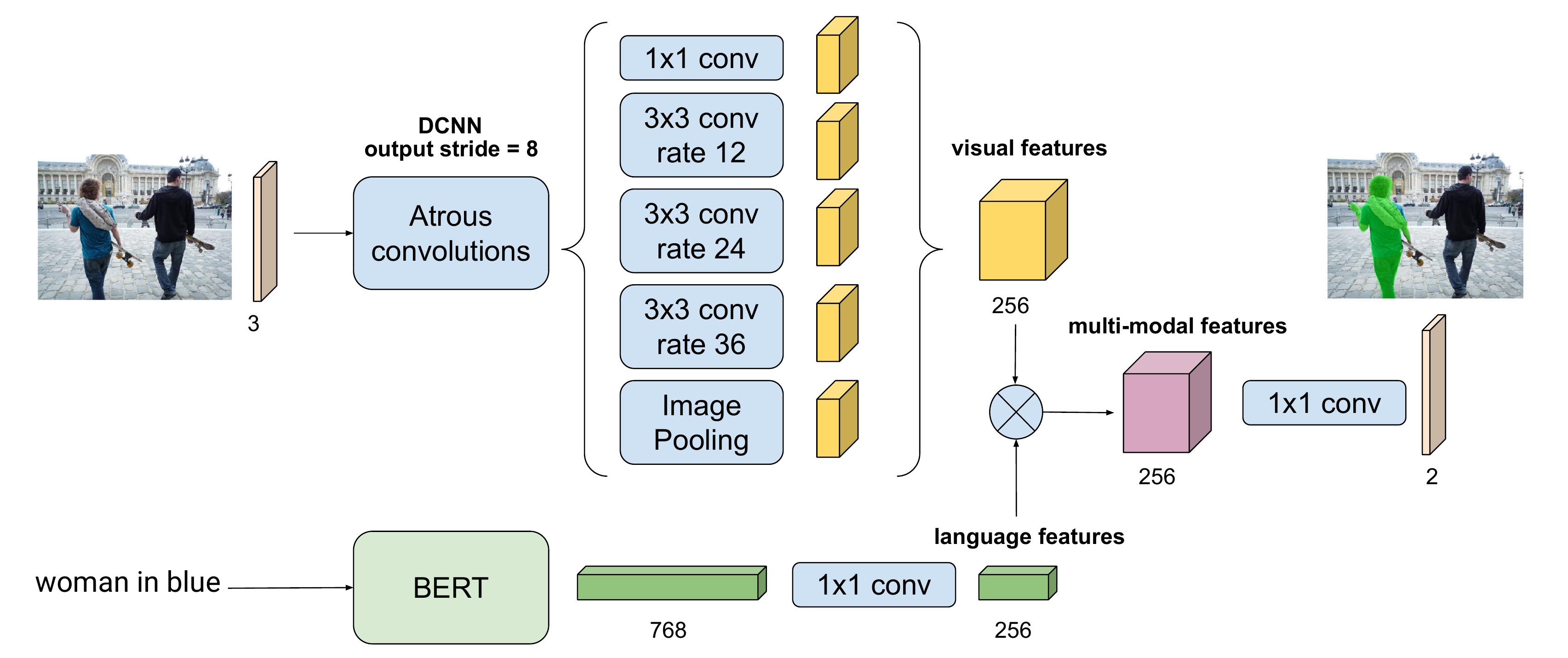}
  \caption{Architecture of our model.}
  \label{fig:single-frame-model}
\end{figure*}

We address the task of language-guided image segmentation with the deep neural network depicted in Figure~\ref{fig:single-frame-model}, that we call RefVOS.
This model operates at the frame level, i.e., treats each frame independently, and is thus applicable for both images and videos.
It uses state of the art visual and language feature extractors, which are combined into a multi-modal embedding decoded to generate a binary mask for the referent.

\noindent
\textbf{Visual Encoder:} To encode the images we rely on DeepLabv3, a network for semantic segmentation based on atrous convolutions~\cite{chen2017rethinking}. We use DeepLabv3 with a ResNet101~\cite{he2016deep} backbone and output stride of 8. The Atrous Spatial Pyramid Pooling~(ASPP) has atrous convolutions with 12, 24 and 36 rates.


\noindent
\textbf{Language Encoder:} In contrast to previous works addressing language-guided image segmentation, we are the first ones to leverage the bidirectional transformer model BERT~\cite{devlin2019bert} as language encoder. For our pipeline, we use BERT to obtain an embedding for the linguistic phrases. First of all we fine-tune BERT 
with the REs of RefCOCO with the masked language modelling~(MLM) loss for one epoch, which consists in randomly masking a percentage of input tokens and then predicting them, following the common fine-tuning procedure for BERT. We then integrate BERT into our pipeline and fine-tune it 
specifically  
towards the language-guided image segmentation task: to this end we tokenize the linguistic phrase and add the [CLS] and [SEP] tokens at the beginning and end of the sentence respectively. BERT produces a 768-dimensional embedding for each input token. We adopt the procedure of Devlin~\etal~\cite{devlin2019bert} and extract the embedding corresponding to the [CLS] input token, i.e.,~the \textit{pooled output}, as it aggregates a representation of the whole sequence. 

\noindent
\textbf{Multi-modal Embedding:} To obtain a multi-modal embedding, the encoded linguistic phrase is first converted to a 256-dimensional embedding with a linear projection and then element-wise multiplied with the visual features extracted by the ASPP from DeepLabv3. 
We noted that the multiplication yielded better performance than addition or concatenation. A convolutional layer then predicts two maps, one for the \textit{foreground} and another for the \textit{background} class. 
We employ the cross entropy loss commonly used for segmentation.

\section{Referring Expression Categorization}
\label{sec:refexps}

We propose a novel categorization for referring expressions (REs), i.e.,~linguistic phrases that allow the identification of an individual object~(the \textsl{referent}) in a discourse or scene. This categorization is adapted to the challenges posed by the VOS task.
We follow the commonly adopted definition of REs put forward by computational linguistics and natural language processing (e.g.,~\cite{reiter1992fast}), and consider a (noun) phrase as a RE if it is an accurate description of the referent, but not of any other object in the current scene.  
Likewise, in Vision \& Language research, visual RE resolution and generation has seen a rise of interest, especially in still images~\cite{cirik2018using,mao2015generation,liu2017referring,yu2016joint,nagaraja16refexpcontext}, and more recently also on videos~\cite{anayurt2019,chen2018real}. 
The task is formulated as, given an instance comprising an image or video with one or multiple objects, and a RE, identify the \textit{referent} that the RE describes by predicting, e.g.,~its bounding box or segmentation mask. 
The difficulty of the task increases with the number of objects appearing in the scene, and the number of objects of the same class. 
Such cases require more complex REs in order to identify the referent. 

In order to make progress on VOS with REs and allow for a systematic comparison of methods, benchmark datasets need to be challenging from both, the visual and linguistic perspective. 
However, for example, most video sequences in the DAVIS-2017 dataset used in Khoreva~\etal~\cite{khoreva2018video} show a single object in the scene or, at most, different objects from different classes. In these cases, the actual challenge is that of predicting accurate object masks for the RE. On the other hand, the existing datasets for VOS with REs do not focus on the particularities that video information provides either, and often use object attributes which can be already captured by a single frame, or are not even true for the whole clip (e.g. the A2D dataset provides phrases for only a few frames per clip). 

Our novel categorization of REs for video objects 
allows the analysis of datasets with respect to the \textit{difficulty} of the REs and the kind of \textit{semantic information} they provide. We apply it to label and analyze existing REs of DAVIS-2017 and A2D. In addition, we use this categorization to extend a subset of the A2D test set with REs which contain semantically varying information to analyze how our model behaves with respect to the different categories.


\subsection{Difficulty and Correctness of Datasets}
We first assess the validity and visual difficulty of a subset of DAVIS-2017 and A2D, by classifying each instance (an object and its RE) into \textit{trivial} or \textit{non-trivial}: if the referent is not the only object of a certain object class in the video 
we consider it \textit{non-trivial}, otherwise \textit{trivial}. 
We further label each phrase according to its linguistic ambiguity and correctness: we mark it as \textit{no RE} if its referent is not the only object in the video which could be described by the phrase, and as \textit{wrong object} if it does not match the referent. 

\noindent
\textbf{Data and Annotation Procedure:}
Annotation was performed on the DAVIS-2017 validation set~($61$ REs provided by \textit{annotator 1}~\cite{khoreva2018video}) in the full video setup~(see Section~\ref{sec:related}), as well as on the subset of the A2D test set which contains at least two annotated objects 
~($856$~instances). 
Each instance contained therein was annotated by one out of four persons~(all co-authors). 
Note that we assume the instances in A2D videos with only a single annotation as \textit{trivial}, and automatically labeled them as such ($439$~instances). 



\begin{figure}[!tbp]
  \centering
  \begin{minipage}[t]{0.23\textwidth}

        \includegraphics[width=\columnwidth]{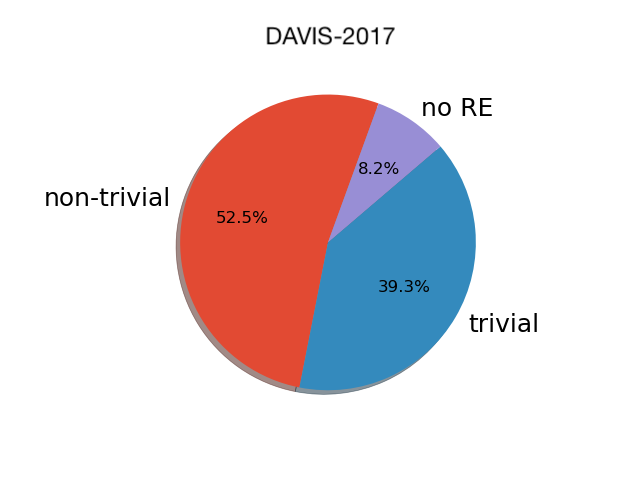}
  \end{minipage}
  \begin{minipage}[t]{0.23\textwidth}
    \includegraphics[width=\columnwidth]{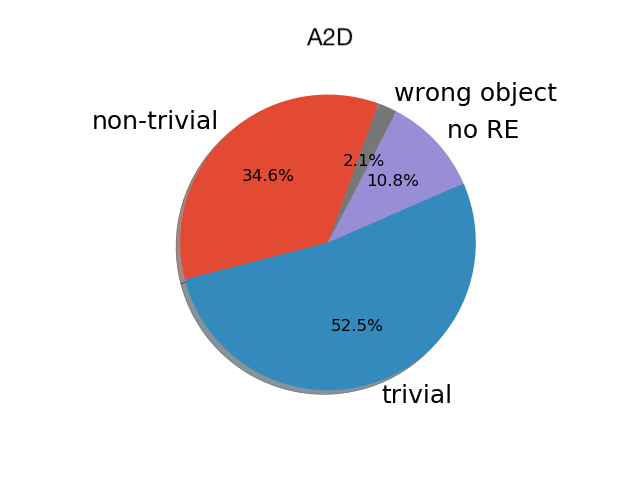}
  \end{minipage}
  \caption{Proportion of expressions in the val set of DAVIS-2017 and the test set of A2D by the difficulty and correctness of the REs.}
  \label{fig:analysis_datasets_difficulty}
\end{figure}

\noindent
\textbf{Results:}
Figure~\ref{fig:analysis_datasets_difficulty} shows the proportion of phrases in the DAVIS-2017 and A2D sets with respect to their difficulty and correctness. 
Despite being collected in a (non-interactive) referential two-player game setup, 
DAVIS-2017 contains a considerable proportion of ambiguous phrases (\textsl{no RE},~$8$\%).  
The proportion in A2D is slighlty higher~($11$\%), but note that A2D was designed to contain descriptive phrases in contrast to unique identifiers (as defined above). 
About~$52\%$ in DAVIS, and $35\%$~in A2D are \textit{non-trivial} phrases, that is, more challenging for language-guided VOS from both, the linguistic and visual perspective, since the object class itself is not sufficient to identify the correct referent. 

\subsection{Semantic Categorization of REs}
Our categorization is inspired by semantic categories of situations and utterances in linguistics \cite{levelt89,germanet1997}, tailored to the situations found in video data. 
Specifically, we analyze the REs with respect to the type of information they express, by assigning them categories assumed to be relevant for reference to objects in visual scenes. 
We focus on information relevant for both, objects in still images and videos, namely the \textsl{category, appearance}, and the \textsl{location} of the referent, and distinguish between information assumed to be more relevant for videos only, namely \textsl{motion} vs. \textsl{static} events.  
If, according to the RE, the referent acts upon other objects in the scene, we distinguish between whether an object is moved by the referent or not (\textsl{obj-motion} vs. \textsl{obj-static}). 
This information may be particularly valuable for models that reason over object interactions. 

\begin{table*}[t]
\centering
\resizebox{0.7\linewidth}{!}{
\begin{tabular}{@{~}l|ll@{~}}
    \toprule
    \textbf{Category}  &\textbf{Q: Does RE tell you about referent~$r$\dots} & \textbf{Example}\\
    \midrule 
     appearance &  how $r$ looks like? & \dots\textit{in a yellow dress}\dots\\
     category &  $r$'s name or category (noun) & \dots \textit{seagull}\dots\\
     location &  where $r$ is located? (rel. to image/other object) & \dots\textit{near tractor}\dots\\
     \midrule
     motion & if $r$ moves or changes its location? & \dots\textit{walking}\dots \\
     obj-motion & if $r$ moves or changes another object's location? & \dots\textit{riding a bike}\dots\\
     static &  what $r$ is doing (if not moving)? & \dots\textit{eating}\dots\\
     obj-static & if $r$ acts on another object (no motion)? & \dots\textit{holding a bike}\dots\\
     \bottomrule
\end{tabular}
}
\caption{The semantic categories used for annotation.}
\label{tab:semcategories}
\end{table*}


(Psycho)linguistic studies have observed a tendency of REs to contain redundant nondiscriminating information, i.e.,~logically more information than required to establish unique reference, arguably because this reduces the effort needed for identification~\cite{hervas2010prevalence, levelt89}. 
In particular the kind~(category) of the object and salient properties such as color have been found to be used redundantly~\cite{rubio2016redundant}. 
To assess whether the phenomenon of redundancy is born out in the video datasets, we additionally label instances as \textit{redundant} or \textit{minimal}. 

\noindent
\textbf{Data and Annotation Procedure:}
We collect annotations for the same $61$~instances of the validation set of DAVIS-2017 as above, and for a subset of the test set of A2D, which we call \textit{A2Dre} henceforth. 
We obtain A2Dre by selecting only instances that were labeled as \textsl{non-trivial}, which are $433$~REs from $190$~videos. We do not use the \textit{trivial} cases as the analysis of such examples is not relevant, as \textit{referents} can be described by using the \textit{category} alone. Each annotator was presented with a RE, a video in which the target object was marked by a bounding box, and a set of questions paraphrasing our categories (see Table~\ref{tab:semcategories}).
Three annotators (all co-authors of the paper) individually labeled all instances of the DAVIS-2017 val set, then jointly discussed their disagreements, and again individually revised their annotations for possible errors or other unclear cases. 
The inter-annotator agreement can be considered substantial for all categories, with Davies \& Fleiss' kappa coefficients \cite{davies1982kappa} between \mbox{$\kappa=.83$} and~$.97$ (except \textsl{obj-static}, \mbox{$\kappa=.35$}, which has only 5 positively labeled instances by at most 2 annotators, and \textsl{category} which obtained perfect agreement). A2Dre was subsequently annotated by the same 3 annotators. 
Our final set of category annotations used for analysis was derived by means of majority voting: for each \textsl{non-trivial} RE, we kept all category labels which were assigned to the RE by at least two annotators.

\paragraph{Results: What kind of information do REs express?}
First of all, we found~$99$\% of the REs for non-trivial instances in A2Dre, and~$66$\% in DAVIS-2017 val ($74$\%~including trivial), respectively, to contain redundant information.
Recall that only the REs in DAVIS-2017 were obtained in a referential setup, thus relatively larger proportion of redundant REs in A2D is not surprising. 

\begin{figure}[!tbp]
  \centering
  \begin{minipage}[t]{0.23\textwidth}

    \includegraphics[width=\columnwidth]{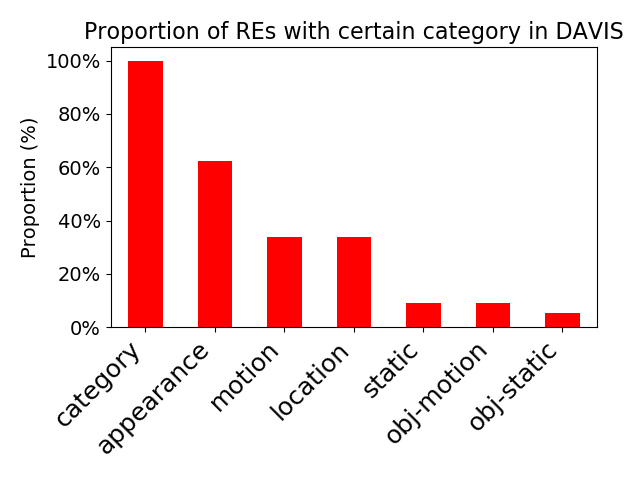}
  \end{minipage}
  \begin{minipage}[t]{0.23\textwidth}

    \includegraphics[width=\columnwidth]{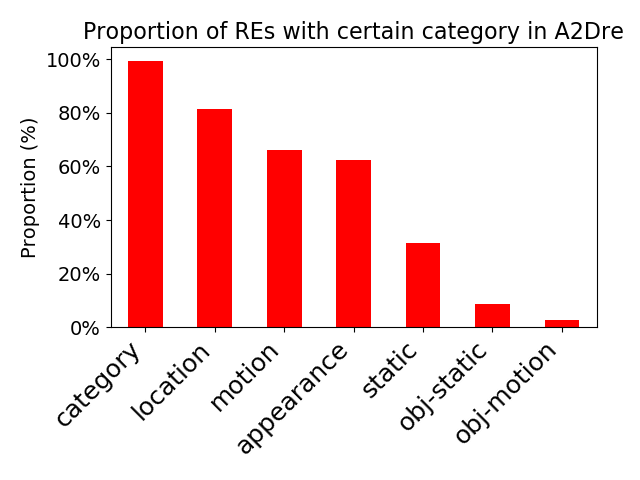}
  \end{minipage}
    \caption{REs in the validation set of DAVIS-2017 and  A2Dre with respect to their categories.}
  \label{fig:analysis_datasets}
\end{figure}

Figure~\ref{fig:analysis_datasets} shows the proportion of instances in the two datasets~(DAVIS-2017 val and A2Dre) that were labeled with the individual categories. 
As expected, the name or \textit{category} of the referent is virtually always expressed. 
The visual properties of the referent, i.e.,~\textit{appearance}, is prominent in both datasets, too (approx.~$60$\%). 
Taken together with their high redundancy ratio, this confirms what has been found in psycholinguistic studies on reference~\cite{levelt89}. 
The remaining categories, however, are rare in both datasets, or are only highly frequent in A2Dre, with \textsl{location} and \textsl{motion} being used in the majority of REs. 
That A2Dre comprises more complex REs than DAVIS-2017 may be not only  due to their collection as descriptive, instead of discrimininative phrases, but also due to the much higher complexity of the video scenes. 
Note that information about referent-object interactions (\textit{obj-static} and \textit{obj-motion}) is neglectable, which illustrates the datasets' limited usefulness for research on reasoning over object interactions~\cite{wang2018NeighbourhoodWR,zhang2019referring,yang2019dynamic}. 
In the experiments we report in Section~\ref{sec:experiments}, we discard these categories, and focus on the remaining categories only, for which we augment the A2Dre dataset.


\subsection{Extending A2D with REs}
\label{sec:data-collection}

As explained above, A2Dre is a subset from the A2D test set including $433$~\textit{non-trivial} REs. 
Due to its highly unbalanced distribution across the $7$~semantic categories~(Figure~\ref{fig:analysis_datasets}), 
we select the $4$~major categories \textsl{appearance, location, motion and static}. 
The four categories have in common that in most cases, for a given referent, a RE can be provided that expresses a certain category, and one that does not. 
We use these categories to augment A2Dre with additional REs, which vary according to the presence or absence of each them. 
Specifically, based on our categorization of the original REs, for each RE~$re$ and category~$C$, we produce an additional RE~$re'$ by modifying $re$ slightly such that it does (or does not) express~$C$. 
For example, if we have the last RE from Figure~\ref{fig:a2d-images}, i.e. \emph{girl in yellow dress standing near the woman}, which could be categorized as \textit{appearance}, \textit{location}, no \textit{motion} and \textit{static}, we produce new REs for each category: \emph{girl standing near the woman} (no \textit{appearance}), \emph{girl in yellow dress standing} (no \textit{location}), \emph{girl in yellow dress walking} (\textit{motion}) and \emph{girl in yellow dress near the woman} (no \textit{static}). 
We do not apply this procedure for \textsl{category}, since it is expressed in almost all REs, and its removal may be difficult in many cases. 
We will refer to this extended dataset as A2Dre+.

\section{Experiments}
\label{sec:experiments}

We report results with our model on two different tasks: \textit{language-guided image segmentation} and \textit{language-guided video object segmentation}. The results for still images are obtained on RefCOCO and RefCOCO+~\cite{yu2016modeling}, while those for video correspond to DAVIS-2017 and A2D. 


\subsection{Language-guided Image Segmentation}
\label{ssec:ExperimentsImage}

The impact of BERT embeddings in our model on both RefCOCO and RefCOCO+ can be assessed in Table~\ref{table:results_refcoco}, compared with a bidirectional LSTM similar to Chen~\etal~\cite{chen2019see} for encoding the linguistic phrase. 
In particular, we average the GloVe embeddings~\cite{pennington2014glove} of each token and concatenate the mean embeddings of the forward and backward pass.
This baseline is compared to two configurations that use BERT. The first fine-tunes BERT for the language-guided image segmentation task, and significantly boosts performance over using GloVe embeddings. The second has an additional step, that consists in first training BERT with the masked language modelling loss with the REs from RefCOCO, as explained in Section~\ref{sec:single-frame-models}, and then fine-tuning BERT on the language-guided image segmentation task~(as in the previous configuration). We see that this configuration brings an additional gain. 

\begin{figure}[!tbp]
  \centering
    \includegraphics[width=0.9\columnwidth]{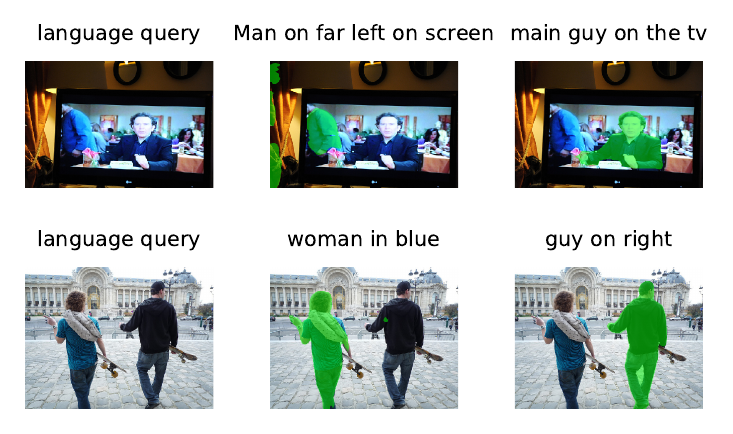}
  \caption{
  Qualitative results obtained on RefCOCO.}
  \label{fig:refcoco_vis}
\end{figure}

Table~\ref{table:results_refcoco} also compares our model with the state of the art on language-guided image segmentation.
STEP~\cite{chen2019see} consists in an iterative model that refines the RE representation to improve the segmentation. Note that the model must be run for each iteration. Our model surpasses STEP~(1-fold) on RefCOCO val and testA, which corresponds to a comparable computational cost, and is still slightly better than STEP~(4-fold). Compared to STEP~(5-fold), the performance of our method is slightly lower. BRINet~\cite{hu2020bi} and CMPC~\cite{huang2020referring} are both superior in terms of performance. However, compared to ours, they are significantly more complex. CMPC is composed of several independent modules and needs to build a relational graph per query. BRINet has a cross-attentional and a bidirectional module to fuse cross-modal features. Both BRINet and CPMC use a Dense-CRF post-processing step~\cite{krahenbuhl2011efficient}. In comparison, our network is simpler and is fully end-to-end. Qualitative results generated with our best model on RefCOCO are depicted in Figure~\ref{fig:refcoco_vis}. We note how our model distinguishes properly the referred instance and generates an accurate mask. We conclude that our approach is competitive with the state of the art for language-guided image segmentation. Hence, \textit{RefVOS} is a valid model for language-guided VOS, and for running an analysis on our RE categorization.

\begin{table}[]
\centering
\resizebox{\linewidth}{!}{
\begin{tabular}{@{}lcccccc@{}}
\toprule
                            & \multicolumn{3}{c}{RefCOCO}                        & \multicolumn{3}{c}{RefCOCO+}        \\
                                                & val            & testA          & testB                   & val     & testA          & testB    \\ \midrule
Ours with Bi-LSTM           & 48.46          & 52.90          & 44.43                   & 35.35        & 40.72               & 28.43        \\
Ours with BERT              & 58.65          & 62.28          & 54.28                   & 42.07      & 46.46 & 34.23        \\
Ours with BERT Pre-train & 59.45          & 63.19          & 54.17                   & 44.71      & 49.73 & 36.17        \\ \midrule
MattNet                 & 56.51          & 62.37          & 51.70  & 46.67   & 52.39          & 40.08         \\
CMSA                      & 58.32          & 60.61          & 55.09  & 43.76   & 47.60          & 37.89          \\
LANG2SEG          & 58.90          & 61.77          & 53.81  & -   & -          & -          \\
STEP~(1-fold)                  & 56.58          & 58.70              & 55.39     & -     & -               & -           \\
STEP~(4-fold)                  & 59.13          & -              & -     & -     & -               & -           \\
STEP~(5-fold)           & 60.04 & 63.46 & 58.97  & 48.18        & 52.33          & 40.41 \\ 
BRINet                  & 61.35 & 63.37 & 59.57  & 48.57        & 52.87          & 42.13 \\ 
CMPC                 & \textbf{61.36} & \textbf{64.53} & \textbf{59.64}  & \textbf{49.56}        & \textbf{53.44}          & \textbf{43.23} \\ 

\bottomrule
\end{tabular}
}
\caption{Overall IoU for RefCOCO and RefCOCO+.}
\label{table:results_refcoco}
\end{table}

\subsection{Language-guided VOS}

Our model is assessed for LVOS on DAVIS-2017 and A2D. In both cases, each video frame is treated separately, so we use the same architecture as in the still image experiments from Section~\ref{ssec:ExperimentsImage}.


Our experiments on the DAVIS-2017 validation set are reported in Table~\ref{table:davis-single-frame}. All models are pre-trained on RefCOCO.
Results are provided with the J\&F metric adopted in the DAVIS-2017 challenge for the two different types of REs collected by Khoreva~\etal~\cite{khoreva2018video} explained in Section~\ref{sec:related}. J\&F is the average between a region-based evaluation measure (J) and a contour-based one (F).
Our experiments indicate that our baseline model trained only with RefCOCO already outperforms the best model by Khoreva~\etal~\cite{khoreva2018video}, despite the latter being fine-tuned on the same DAVIS-2017 dataset (+Ft DAVIS segms.).
The difference increases when our model is fine-tuned with the segmentations provided in the training set, but freezing the language encoder.
This is the configuration comparable to Khoreva~\etal~\cite{khoreva2018video} in terms of training data, and brings gains of 2.7 and 4.9 points for the \textit{first frame} and \textit{full video} REs, respectively.
Finally, we also fine-tune the BERT language encoder, obtaining a significant extra gain in performance.
We want to highlight that our frame-based model does not rely on any post-processing to add temporal coherence, or optical flow, in contrast to Khoreva~\etal~\cite{khoreva2018video}, so our method may be more efficient computationally.  We also compare our model to URVOS~\cite{seourvos}, a concurrent work to ours. RefVOS performs slightly better when trained with the same amount of annotated data. Qualitative results for full video REs are shown in Figure~\ref{fig:davis_best_full_video}. When the multiple objects belong to different categories, the model works produces accurate masks from the language query, whereas it is more challenging to properly segment the referent in cases where there are multiple instances of the same class in the sequence~(3rd row). 
The fine-tuning is done with the \textit{full video} REs, and the REs shown in Figure~\ref{fig:davis_best_full_video} are of the same kind. We note how the referred object is in general identified and properly segmented.

\begin{table}[]
\centering
\resizebox{\linewidth}{!}{
\begin{tabular}{@{}cccccc@{}}
\toprule
\multirow{2}{*}{Model}    & +Ft DAVIS  & \multicolumn{2}{c}{+Ft DAVIS REs}  & \multicolumn{2}{c}{J\&F}          \\
                           & segms.        &   1st frame              & full video                        & 1st frame       & full video    \\ 
\midrule
Khoreva~\etal~\cite{khoreva2018video}                &  $\checkmark$     &                          &                                   & 39.3              &  37.1           \\
URVOS~\cite{seourvos}                &  $\checkmark$     &  $\checkmark$                        &                                   & 44.1              &  -          \\
\midrule
                                  &                    &                          &                                   & 39.8              &  40.8         \\
     RefVOS                         &  $\checkmark$      &                          &                                 & 42.0              & 42.0          \\
                                  &  $\checkmark$      &  $\checkmark$            &                                   & \textbf{44.5}              & \textbf{45.1} \\
                                  &  $\checkmark$      &                          & $\checkmark$                      & 42.7              & \textbf{45.1}  \\ 
\bottomrule
\end{tabular}
}
\caption{J\&F on DAVIS-2017 validaton set.}
\label{table:davis-single-frame}
\end{table}

\begin{table}[]
\centering
\small
\begin{tabular}{@{}lc@{}ccc@{}}
\toprule
 & \multicolumn{2}{c}{Prec} &  \multicolumn{2}{c}{IoU} \\ 
 & @0.5 & @0.9  & Overall     & Mean        \\ 
\midrule
Gavrilyuk~\etal~\cite{gavrilyuk2018actor}                & 50.0       & 0.4      & 55.1           & 42.6           \\
Wang~\etal~\cite{wang2019asymmetric}                & 55.7      & 2.0    & 60.1           & 49.0                \\        
\midrule
RefVOS with A2D & 49.5      & 6.4         & 59.9          & 43.0          \\
RefVOS with RefCOCO   & 27.9       & 3.4        & 41.4          & 25.6                \\
+ finetuned on A2D              & \textbf{57.8}  & \textbf{9.3}  & \textbf{67.2} & \textbf{49.7} \\
\bottomrule
\end{tabular}
\caption{\textit{Precision}, \textit{overall IoU} and \textit{mean IoU} on A2D.}
\label{table:a2a-single-frame}
\end{table}

\begin{figure*}[!tbp]

  \centering

    
        \begin{minipage}[t]{0.42075\textwidth}

    \includegraphics[width=\columnwidth]{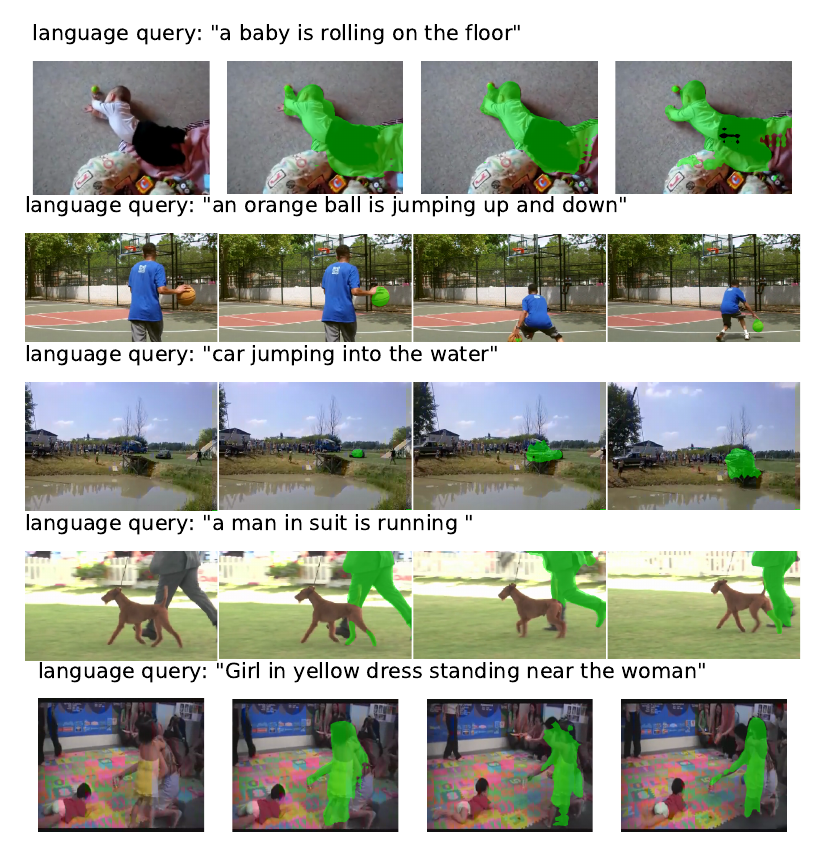}
    \label{fig:refcoco_best_testA}
  \end{minipage}
        \begin{minipage}[t]{0.3213\textwidth}

    \includegraphics[width=\columnwidth]{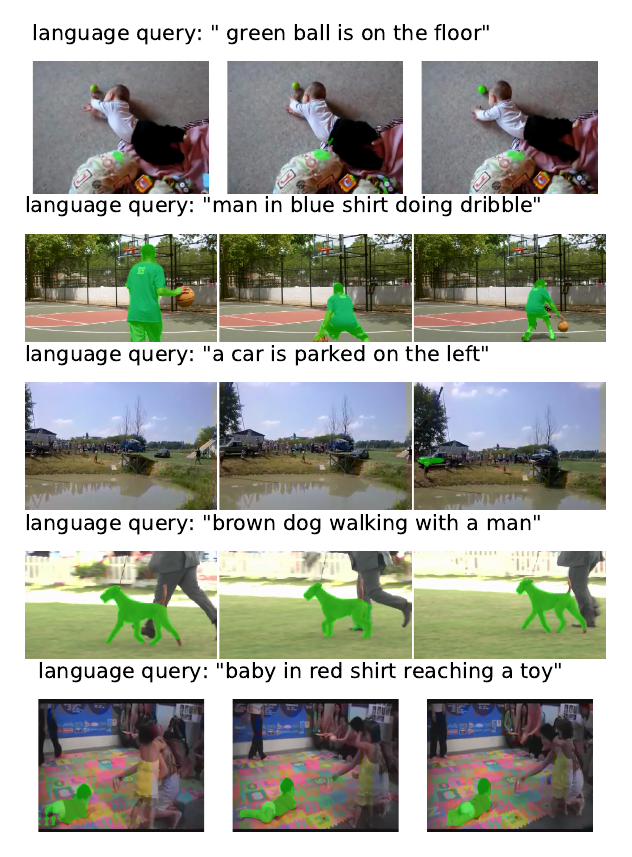}
   
  \end{minipage}
  \caption{Video sequences for A2D with language queries and the results of our model. The first column shows a reference frame, the second to fourth columns depict the masks produced by our model when given the language query shown on top. Finally, the fifth to seventh columns show the results for the language query shown on top of these columns, which refers to another object of the video sequence.}
  \label{fig:a2d-images}
\end{figure*}

The results for A2D are shown in Table~\ref{table:a2a-single-frame}, using the metrics that allow us a comparison with previous works~\cite{gavrilyuk2018actor,wang2019asymmetric}. 
Our model trained only with A2D already outperforms Gavrilyuk~\etal~\cite{gavrilyuk2018actor} in \textit{Precision} at a high threshold and at the \textit{Overall} and \textit{Mean Intersection Over Union~(IoU)}.
Moreover, our model significantly increases its performance when it is first trained on RefCOCO and later fine-tuned on A2D, both its visual and language branches. In this setup, it achieves state of the art results in all metrics by significant margins. Note that both Gavrilyuk~\etal~\cite{gavrilyuk2018actor} and Wang~\etal~\cite{wang2019asymmetric} leverage an encoder pre-trained on the Kinetics dataset, which includes 650,000 video clips~\cite{carreira2017quo}. Hence, these models see a large amount of annotated data for action recognition in videos.
We also want to stress our high \textit{Precision} values at high thresholds, which indicates that our model is able to produce very accurate masks. 
Visualizations with our model are illustrated in Figure~\ref{fig:a2d-images}.

In conclusion, RefVOS is state of the art for DAVIS-2017 and A2D on the LVOS task, although it is a frame-based model. This motivates the analysis of our model when tested with different types of REs, based on the categorization and difficulty analysis proposed in Section \ref{sec:refexps}.

\noindent
\textbf{REs Analysis:} Firstly, we analyze the performance on \textit{trivial} and \textit{non-trivial} linguistic phrases for both the A2D test and DAVIS-2017 validation sets. 
The \textit{mean IoU} per referent obtained for \textit{trivial} and \textit{non-trivial} for DAVIS-2017 is 48.7 \textit{vs.} 46.2, and for A2D is 53.9 \textit{vs.} 33.2.
We observe that the performance is worse for the \textit{non-trivial} cases for both datasets as expected, with a major drop on A2D. 

Secondly, we study the effect of RE categories in relation to the performance of RefVOS. 
The A2Dre+ dataset described in Section~\ref{sec:data-collection} allows us to have the same number of referents for all major categories: \textit{appearance, location, motion} and \textit{static}.
Each of our referents is annotated with highly similar REs~(two for each category) and thus are directly comparable.
In contrast, Khoreva~\etal~\cite{khoreva2018video} split the videos into two different subsets with non-comparable referents.
Table~\ref{table:add-rm-attr} compares the performance of RefVOS depending whether each of the categories is present in the RE.
The results show that the presence of \textit{appearance} and \textit{location} categories yields significantly higher results compared to their absence. 
We also observe a drop in performance when the \textit{static} category is present, which indicates that the model struggles at identifying a referent based on static actions such as \textit{holding, sitting, eating}. 
In contrast, the presence or absence of the \textit{motion} category does not affect the performance, which actually means that the model is unable to benefit from this type of REs.

\begin{table}[]
\centering
\small
\centering
\begin{tabular}{@{}c@{\hspace{3.5pt}}c@{\hspace{3.5pt}}c@{\hspace{3.5pt}}c@{\hspace{3.5pt}}c@{~}c@{\hspace{3.5pt}}c@{~}c@{}}
\toprule
+App & -App & +Loc & -Loc & +Motion & -Motion & +Static & -Static \\
\midrule
\textbf{33.90}      & 30.15      &  \textbf{34.15}      & 30.78     &  35.58        & \textbf{35.60}        & 34.28         & \textbf{36.21}         \\ \bottomrule
\end{tabular}
\caption{Effect of the presence of categories in REs.}
\label{table:add-rm-attr}
\end{table}

\begin{table}[]
\centering
\resizebox{\linewidth}{!}{
\begin{tabular}{@{}lcccccccc@{}}

\toprule
 & \multicolumn{3}{c}{Overall IoU}  & \multicolumn{3}{c}{Mean IoU}  \\ 
                      & Trivial & Non-Trivial & All  & Trivial & Non-Trivial & All       \\ \midrule

Generic & 45.6 & 18.1 & 41.6 & 34.6    & 10.0       & 29.6          \\
Only Actor & 65.6  & 34.8 & 60.8 & 51.5            & 22.8               & 45.7          \\
Only Action & 56.3 & 30.7 & 52.6  & 43.0          & 18.5                & 38.0                \\
Actor + Action & 66.6 & 37.3 & 62.2 & 51.3     & 24.8         & 45.9     \\ 
Full phrase & \textbf{70.2} & \textbf{47.5} & \textbf{67.2} & \textbf{53.9}      &  \textbf{33.2}      & \textbf{49.7} \\
\bottomrule
\end{tabular}
}
\caption{\textit{Overall} and \textit{Mean IoU} on A2D for different levels of information in REs.}
\label{table:a2a-ablation-res}
\end{table}



Finally, in Table~\ref{table:a2a-ablation-res} we study the effect of feeding the model with only the \textit{actor}, the \textit{action}, or the \textit{actor and action}, without formulating any RE, for all the test set of A2D.
These \textit{actor} and \textit{action} terms are obtained from 
the dataset collected by Gavrilyuk~\etal~\cite{gavrilyuk2018actor}. 
In most cases these expressions are not REs as they do not unambiguously describe the referent in the video.
Additionally, we consider a generic phrase \textit{thing}. We distinguish between the \textit{trivial} and \textit{non-trivial} cases.
Results show that RefVOS works significantly better when the \textit{actor} is provided than when the \textit{action} is. Furthermore, performance improves when using both. Finally, having the full linguistic phrase is still the best model. Remarkably, our configuration with \textit{actor and action} reaches higher \textit{Overall IoU} than previous works that use complete linguistic phrases~(see Table~\ref{table:a2a-single-frame}). 
Notice that using the full phrase improves performance especially for the \textit{non-trivial} cases, as these require complete linguistic expressions to identify the \textit{referent}. We also want to stress that the aggregated performance, i.e., considering \textit{all} cases, is dominated by the performance of the \textit{trivial} ones, as they represent most of the dataset. 

\section{Conclusions}
\label{sec:conclusions}



This work studies the difficulty of REs from benchmarks on LVOS, and proposes seven semantic categories to analyze the nature of such REs. We introduce RefVOS, a novel model that is competitive for language-guided image segmentation, and state of the art for language-guided VOS. However, our analysis shows that benchmarks are mainly composed of trivial cases, in which referents can be identified with simple phrases. This indicates that the reported metrics for the task may be misleading. Thus, we focus on the non-trivial cases. We extend A2D with new REs with diverse semantic categories for non-trivial cases, and test our model with them, which reveals that it struggles at exploiting motion and static events, and that it mainly benefits from REs based on appearance and location. We reckon that future research on LVOS should focus on non-trivial cases describing motion and events, as they present a challenge for language grounding on videos. Concurrent to our work, Seo~\etal~\cite{seourvos} collected Refer-Youtube-VOS, a large-scale benchmark for language-guided video object segmentation built on top of Youtube-VOS~\cite{xu2018youtube}. We believe that, as future work, our categorization for REs could be used to classify the provided language expressions by this benchmark. Thus, models could be evaluated based on the non-trivial cases and the different categories in order to analyze which REs are more challenging when using a large-scale dataset.

{\small
\bibliographystyle{ieee_fullname}
\bibliography{egpaper}

\begin{thebibliography}{10}\itemsep=-1pt

\bibitem{anayurt2019}
Hazan Anayurt, Sezai~Artun Ozyegin, Ulfet Cetin, Utku Aktas, and Sinan Kalkan.
\newblock Searching for ambiguous objects in videos using relational referring
  expressions.
\newblock In {\em Proceedings of the British Machine Vision Conference (BMVC)},
  2019.

\bibitem{carreira2017quo}
Joao Carreira and Andrew Zisserman.
\newblock Quo vadis, action recognition? a new model and the kinetics dataset.
\newblock In {\em proceedings of the IEEE Conference on Computer Vision and
  Pattern Recognition}, pages 6299--6308, 2017.

\bibitem{chen2019see}
Ding-Jie Chen, Songhao Jia, Yi-Chen Lo, Hwann-Tzong Chen, and Tyng-Luh Liu.
\newblock See-through-text grouping for referring image segmentation.
\newblock In {\em Proceedings of the IEEE International Conference on Computer
  Vision}, pages 7454--7463, 2019.

\bibitem{chen2017rethinking}
Liang-Chieh Chen, George Papandreou, Florian Schroff, and Hartwig Adam.
\newblock Rethinking atrous convolution for semantic image segmentation.
\newblock {\em arXiv preprint arXiv:1706.05587}, 2017.

\bibitem{chen2018real}
Xinpeng Chen, Lin Ma, Jingyuan Chen, Zequn Jie, Wei Liu, and Jiebo Luo.
\newblock Real-time referring expression comprehension by single-stage
  grounding network.
\newblock {\em arXiv preprint arXiv:1812.03426}, 2018.

\bibitem{cirik2018using}
Volkan Cirik, Taylor Berg-Kirkpatrick, and Louis-Philippe Morency.
\newblock Using syntax to ground referring expressions in natural images.
\newblock {\em AAAI}, 2018.

\bibitem{davies1982kappa}
Mark Davies and Joseph~L. Fleiss.
\newblock {Measuring Agreement for Multinomial Data}.
\newblock {\em Biometrics}, 38(4):1047--1051, 1982.

\bibitem{devlin2019bert}
Jacob Devlin, Ming-Wei Chang, Kenton Lee, and Kristina Toutanova.
\newblock Bert: Pre-training of deep bidirectional transformers for language
  understanding.
\newblock In {\em Proceedings of the 2019 Conference of the North American
  Chapter of the Association for Computational Linguistics}, pages 4171--4186,
  2019.

\bibitem{feng2020real}
Qi Feng, Vitaly Ablavsky, Qinxun Bai, Guorong Li, and Stan Sclaroff.
\newblock Real-time visual object tracking with natural language description.
\newblock In {\em The IEEE Winter Conference on Applications of Computer
  Vision}, pages 700--709, 2020.

\bibitem{gavrilyuk2018actor}
Kirill Gavrilyuk, Amir Ghodrati, Zhenyang Li, and Cees~GM Snoek.
\newblock Actor and action video segmentation from a sentence.
\newblock In {\em Proceedings of the IEEE Conference on Computer Vision and
  Pattern Recognition}, pages 5958--5966, 2018.

\bibitem{germanet1997}
Birgit Hamp and Helmut Feldweg.
\newblock {G}erma{N}et - a lexical-semantic net for {G}erman.
\newblock In {\em Automatic Information Extraction and Building of Lexical
  Semantic Resources for {NLP} Applications}, 1997.

\bibitem{he2016deep}
Kaiming He, Xiangyu Zhang, Shaoqing Ren, and Jian Sun.
\newblock Deep residual learning for image recognition.
\newblock In {\em Proceedings of the IEEE conference on computer vision and
  pattern recognition}, pages 770--778, 2016.

\bibitem{hervas2010prevalence}
Raquel Herv{\'a}s and Mark Finlayson.
\newblock The prevalence of descriptive referring expressions in news and
  narrative.
\newblock In {\em Proceedings of the {ACL} 2010 Conference Short Papers}, pages
  49--54, Uppsala, Sweden, July 2010. Association for Computational
  Linguistics.

\bibitem{hu2016segmentation}
Ronghang Hu, Marcus Rohrbach, and Trevor Darrell.
\newblock Segmentation from natural language expressions.
\newblock In {\em European Conference on Computer Vision}, pages 108--124.
  Springer, 2016.

\bibitem{hu2020bi}
Zhiwei Hu, Guang Feng, Jiayu Sun, Lihe Zhang, and Huchuan Lu.
\newblock Bi-directional relationship inferring network for referring image
  segmentation.
\newblock In {\em Proceedings of the IEEE/CVF Conference on Computer Vision and
  Pattern Recognition}, pages 4424--4433, 2020.

\bibitem{huang2020referring}
Shaofei Huang, Tianrui Hui, Si Liu, Guanbin Li, Yunchao Wei, Jizhong Han, Luoqi
  Liu, and Bo Li.
\newblock Referring image segmentation via cross-modal progressive
  comprehension.
\newblock In {\em Proceedings of the IEEE/CVF Conference on Computer Vision and
  Pattern Recognition}, pages 10488--10497, 2020.

\bibitem{johnson2017clevr}
Justin Johnson, Bharath Hariharan, Laurens van~der Maaten, Li Fei-Fei, C
  Lawrence~Zitnick, and Ross Girshick.
\newblock Clevr: A diagnostic dataset for compositional language and elementary
  visual reasoning.
\newblock In {\em Proceedings of the IEEE Conference on Computer Vision and
  Pattern Recognition}, pages 2901--2910, 2017.

\bibitem{kazemzadeh2014referitgame}
Sahar Kazemzadeh, Vicente Ordonez, Mark Matten, and Tamara Berg.
\newblock {ReferItGame: Referring to Objects in Photographs of Natural Scenes}.
\newblock In {\em Proceedings of the 2014 Conference on Empirical Methods in
  Natural Language Processing ({EMNLP})}, pages 787--798, 2014.

\bibitem{khoreva2018video}
Anna Khoreva, Anna Rohrbach, and Bernt Schiele.
\newblock Video object segmentation with language referring expressions.
\newblock In {\em Asian Conference on Computer Vision}, pages 123--141.
  Springer, 2018.

\bibitem{krahenbuhl2011efficient}
Philipp Kr{\"a}henb{\"u}hl and Vladlen Koltun.
\newblock Efficient inference in fully connected crfs with gaussian edge
  potentials.
\newblock In {\em Advances in neural information processing systems}, pages
  109--117, 2011.

\bibitem{levelt89}
William J.~M. Levelt.
\newblock {\em Speaking: From Intention to Articulation}.
\newblock MIT Press, Cambridge, MA, 1989.

\bibitem{li2017tracking}
Zhenyang Li, Ran Tao, Efstratios Gavves, Cees~GM Snoek, and Arnold~WM
  Smeulders.
\newblock Tracking by natural language specification.
\newblock In {\em Proceedings of the IEEE Conference on Computer Vision and
  Pattern Recognition}, pages 6495--6503, 2017.

\bibitem{lin2014mscoco}
Tsung-Yi Lin, Michael Maire, Serge Belongie, James Hays, Pietro Perona, Deva
  Ramanan, Piotr Doll{\'a}r, and C.~Lawrence Zitnick.
\newblock Microsoft coco: Common objects in context.
\newblock In David Fleet, Tomas Pajdla, Bernt Schiele, and Tinne Tuytelaars,
  editors, {\em Computer Vision -- ECCV 2014}, pages 740--755, Cham, 2014.
  Springer International Publishing.

\bibitem{liu2017recurrent}
Chenxi Liu, Zhe Lin, Xiaohui Shen, Jimei Yang, Xin Lu, and Alan Yuille.
\newblock Recurrent multimodal interaction for referring image segmentation.
\newblock In {\em Proceedings of the IEEE International Conference on Computer
  Vision}, pages 1271--1280, 2017.

\bibitem{liu2019learning}
Daqing Liu, Hanwang Zhang, Feng Wu, and Zheng-Jun Zha.
\newblock Learning to assemble neural module tree networks for visual
  grounding.
\newblock In {\em Proceedings of the IEEE International Conference on Computer
  Vision}, pages 4673--4682, 2019.

\bibitem{liu2017referring}
Jingyu Liu, Liang Wang, and Ming{-}Hsuan Yang.
\newblock Referring expression generation and comprehension via attributes.
\newblock In {\em {IEEE} International Conference on Computer Vision, {ICCV}
  2017, Venice, Italy, October 22-29, 2017}, pages 4866--4874. {IEEE} Computer
  Society, 2017.

\bibitem{mao2015generation}
Junhua Mao, Jonathan Huang, Alexander Toshev, Oana Camburu, Alan~L Yuille, and
  Kevin Murphy.
\newblock Generation and comprehension of unambiguous object descriptions.
\newblock In {\em Proceedings of the IEEE conference on computer vision and
  pattern recognition}, pages 11--20, 2016.

\bibitem{nagaraja16refexpcontext}
Varun~K. Nagaraja, Vlad~I. Morariu, and Larry~S. Davis.
\newblock Modeling context between objects for referring expression
  understanding.
\newblock In {\em European Conference on Computer Vision (ECCV)}, 2016.

\bibitem{pennington2014glove}
Jeffrey Pennington, Richard Socher, and Christopher~D Manning.
\newblock Glove: Global vectors for word representation.
\newblock In {\em Proceedings of the 2014 conference on empirical methods in
  natural language processing (EMNLP)}, pages 1532--1543, 2014.

\bibitem{perazzi2016benchmark}
Federico Perazzi, Jordi Pont-Tuset, Brian McWilliams, Luc Van~Gool, Markus
  Gross, and Alexander Sorkine-Hornung.
\newblock A benchmark dataset and evaluation methodology for video object
  segmentation.
\newblock In {\em Proceedings of the IEEE Conference on Computer Vision and
  Pattern Recognition}, pages 724--732, 2016.

\bibitem{pont20172017}
Jordi Pont-Tuset, Federico Perazzi, Sergi Caelles, Pablo Arbel{\'a}ez, Alex
  Sorkine-Hornung, and Luc Van~Gool.
\newblock The 2017 davis challenge on video object segmentation.
\newblock {\em arXiv preprint arXiv:1704.00675}, 2017.

\bibitem{reiter1992fast}
Ehud Reiter and Robert Dale.
\newblock A fast algorithm for the generation of referring expressions.
\newblock In {\em {COLING} 1992 Volume 1: The 15th {I}nternational {C}onference
  on {C}omputational {L}inguistics}, 1992.

\bibitem{rubio2016redundant}
Paula Rubio-Fernández.
\newblock How redundant are redundant color adjectives? an efficiency-based
  analysis of color overspecification.
\newblock {\em Frontiers in Psychology}, 7:153, 2016.

\bibitem{seourvos}
Seonguk Seo, Joon-Young Lee, and Bohyung Han.
\newblock Urvos: Unified referring video object segmentation network with a
  large-scale benchmark.
\newblock In {\em Proceedings of the European Conference on Computer Vision
  (ECCV)}, 2020.

\bibitem{simonyan2014very}
Karen Simonyan and Andrew Zisserman.
\newblock Very deep convolutional networks for large-scale image recognition.
\newblock In {\em International Conference on Learning Representations (ICLR)},
  2015.

\bibitem{wang2019asymmetric}
Hao Wang, Cheng Deng, Junchi Yan, and Dacheng Tao.
\newblock Asymmetric cross-guided attention network for actor and action video
  segmentation from natural language query.
\newblock In {\em Proceedings of the IEEE International Conference on Computer
  Vision}, pages 3939--3948, 2019.

\bibitem{wang2018NeighbourhoodWR}
Peng Wang, Qi Wu, Jiewei Cao, Chunhua Shen, Lianli Gao, and Anton van~den
  Hengel.
\newblock Neighbourhood watch: Referring expression comprehension via
  language-guided graph attention networks.
\newblock {\em 2019 IEEE/CVF Conference on Computer Vision and Pattern
  Recognition (CVPR)}, pages 1960--1968, 2018.

\bibitem{xu2015can}
Chenliang Xu, Shao-Hang Hsieh, Caiming Xiong, and Jason~J Corso.
\newblock Can humans fly? action understanding with multiple classes of actors.
\newblock In {\em Proceedings of the IEEE Conference on Computer Vision and
  Pattern Recognition}, pages 2264--2273, 2015.

\bibitem{xu2018youtube}
Ning Xu, Linjie Yang, Yuchen Fan, Jianchao Yang, Dingcheng Yue, Yuchen Liang,
  Brian Price, Scott Cohen, and Thomas Huang.
\newblock Youtube-vos: Sequence-to-sequence video object segmentation.
\newblock In {\em Proceedings of the European Conference on Computer Vision
  (ECCV)}, pages 585--601, 2018.

\bibitem{yang2019dynamic}
Sibei Yang, Guanbin Li, and Yizhou Yu.
\newblock Dynamic graph attention for referring expression comprehension.
\newblock {\em 2019 IEEE/CVF International Conference on Computer Vision
  (ICCV)}, pages 4643--4652, 2019.

\bibitem{ye2019cross}
Linwei Ye, Mrigank Rochan, Zhi Liu, and Yang Wang.
\newblock Cross-modal self-attention network for referring image segmentation.
\newblock In {\em Proceedings of the IEEE Conference on Computer Vision and
  Pattern Recognition}, pages 10502--10511, 2019.

\bibitem{yu2018mattnet}
Licheng Yu, Zhe Lin, Xiaohui Shen, Jimei Yang, Xin Lu, Mohit Bansal, and
  Tamara~L Berg.
\newblock Mattnet: Modular attention network for referring expression
  comprehension.
\newblock In {\em Proceedings of the IEEE Conference on Computer Vision and
  Pattern Recognition}, pages 1307--1315, 2018.

\bibitem{yu2016modeling}
Licheng Yu, Patrick Poirson, Shan Yang, Alexander~C Berg, and Tamara~L Berg.
\newblock Modeling context in referring expressions.
\newblock In {\em European Conference on Computer Vision}, pages 69--85.
  Springer, 2016.

\bibitem{yu2016joint}
Licheng Yu, Hao Tan, Mohit Bansal, and Tamara~L Berg.
\newblock A joint speaker-listener-reinforcer model for referring expressions.
\newblock In {\em Proceedings of the IEEE Conference on Computer Vision and
  Pattern Recognition}, pages 7282--7290, 2017.

\bibitem{zhang2019referring}
Chao Zhang, Weiming Li, Wanli Ouyang, Qiang Wang, Woo-Shik Kim, and Sunghoon
  Hong.
\newblock {Referring Expression Comprehension with Semantic Visual Relationship
  and Word Mapping}.
\newblock In {\em Proceedings of the 27th ACM International Conference on
  Multimedia}, MM ’19, page 1258–1266, Nice, France, 2019. Association for
  Computing Machinery.

\end{thebibliography}
}

\end{document}